\documentclass{article}

\PassOptionsToPackage{numbers, compress}{natbib}

\usepackage[preprint]{neurips_2026}

\usepackage[utf8]{inputenc} % allow utf-8 input
\usepackage[T1]{fontenc}    % use 8-bit T1 fonts
\usepackage{hyperref}       % hyperlinks
\usepackage{url}            % simple URL typesetting
\usepackage{booktabs}       % professional-quality tables
\usepackage{amsfonts}       % blackboard math symbols
\usepackage{nicefrac}       % compact symbols for 1/2, etc.
\usepackage{microtype}      % microtypography
\usepackage{xcolor}         % colors
\usepackage{amsmath}
\usepackage{multirow}
\usepackage{graphicx}
\usepackage{wrapfig}

\title{SurgVista: Long-Horizon Surgical World Modeling with Plausible Instrument-Tissue Dynamics}

\author{%
{\bfseries Wentao Pan$^{1}$ \hspace{2.2em} Wuyang Li$^{2}$ \hspace{2.2em} Shengyuan Liu$^{1}$}\\[0.1cm]
{\bfseries Xinyu Liu$^{3}$ \hspace{2.2em} Hengyu Liu$^{1}$ \hspace{2.2em} Yixuan Yuan$^{1}$\thanks{Corresponding Author}}\\[0.24cm]
{\normalfont $^{1}$The Chinese University of Hong Kong \quad $^{2}$EPFL \quad $^{3}$Imperial College London}\\
}

\begin{document}

\maketitle
\begin{abstract}
Scaling robot policy learning for autonomous surgery is challenging, as expert demonstrations are expensive and in vivo exploration poses substantial safety risks.
Surgical world models address this by generating realistic, action-conditioned future frames from an initial observation, but existing methods exhibit two persistent failure modes: \emph{spatial interaction incoherence}, where visible instrument contact fails to induce spatially consistent tissue deformation, and \emph{temporal fidelity collapse}, where prediction errors compound across autoregressive rollouts and progressively corrupt visual quality.
We present \textbf{SurgVista}, a surgical world model that mitigates both failures through two training recipes.
\emph{Deformation Consistency Regularization} extracts scene-point trajectories from training videos and enforces cross-frame coherence through latent contrastive learning, strengthening physically consistent instrument-tissue dynamics.
\emph{Drift Adaptation Training} mitigates long-horizon drift by perturbing conditioning frames with online prediction residuals and photometric augmentations calibrated to long-horizon drift statistics, sustaining visual fidelity over extended rollouts.
To enable rigorous evaluation, we further introduce \textbf{SurgWorld-Bench}, featuring diverse procedure types, long-range rollouts, and decoupled metrics for instrument-motion accuracy and tissue-response fidelity.
Extensive experiments show that SurgVista consistently outperforms state-of-the-art methods across visual quality, temporal consistency, and interaction fidelity, with gains widening as the prediction horizon grows.
\end{abstract}
\section{Introduction}
\begin{figure}[t]
\centering
\includegraphics[width=\linewidth]{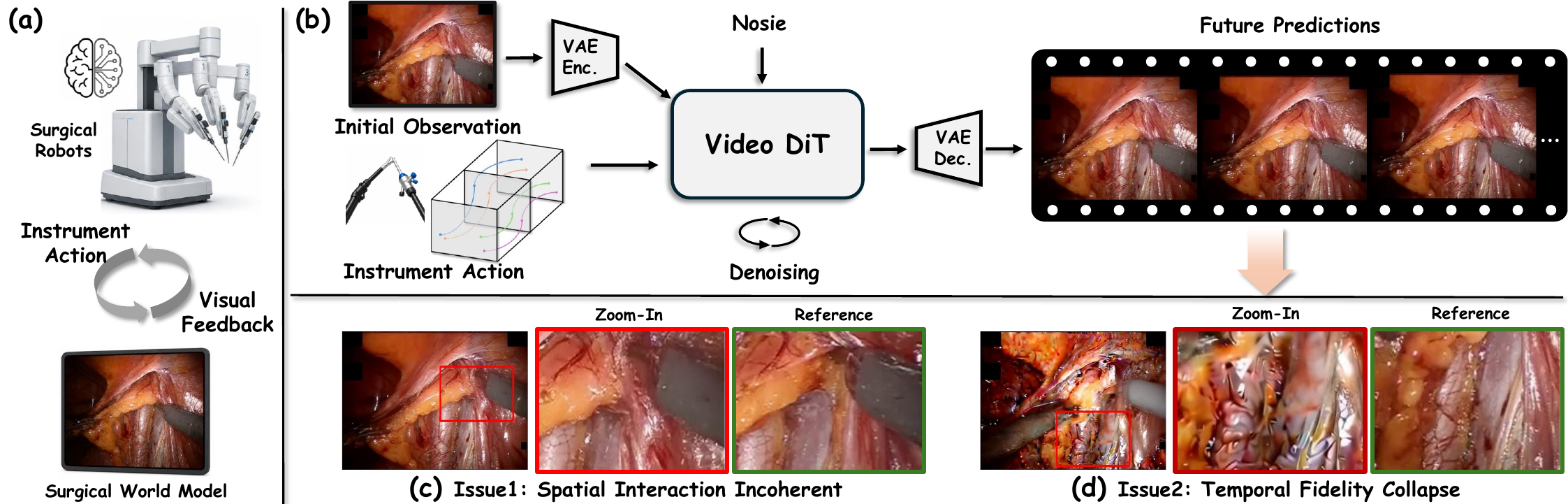}
\vspace{-20pt}
\caption{\textbf{Surgical world modeling.}
(a) Simulating scene evolution for safe policy learning.
(b) Predicting future frames from an initial frame and instrument actions.
\textbf{Two key challenges:}
(c) Tissue fails to deform under instrument contact.
(d) Visual quality degrades over long-horizon rollouts.}
\vspace{-15pt}
\label{fig:teaser}
\end{figure}

Autonomous surgery holds transformative potential to reduce surgeon workload, improve the quality of operative care, and expand access to high-quality surgical treatment~\cite{schmidgall2024general,ciuti2025robotic}. However, scaling robot policy learning through real-world interaction remains difficult, as expert demonstrations are expensive to acquire and in vivo exploration poses substantial safety risks. To address this, Surgical World Models (SWMs)~\cite{yang2025medical}\footnote{Unlike generic-domain world models~\cite{ha2018world,sun2025worldplay,he2025matrix}, SWMs must model realistic instrument-tissue feedback under precise instrument control. This requires capturing contact-driven tissue deformation under challenging endoscopic imaging conditions, including low-light or non-uniform illumination, specular reflections, smoke, and occlusions.} offer a safe and scalable alternative for learning in simulated surgical environments (Fig.~\ref{fig:teaser}a). {\emph{Conditioned on an initial observation and planned instrument actions, such as tool trajectories, an SWM rolls out realistic, action-consistent future representations, e.g., videos}}, enabling policy development and evaluation without patient risk (Fig.~\ref{fig:teaser}b).

Recent SWM works mainly focus on two complementary directions. To improve visual realism, several works~\cite{he2025surgworld,turkcan2025towards,rapuri2026saw} transfer foundational video DiTs~\cite{yang2024cogvideox,wu2025hunyuanvideo} to the surgical domain, producing realistic endoscopic videos under instruction-based control. To improve action controllability, the control signal has progressed from coarse textual descriptions~\cite{he2025surgworld,turkcan2025towards} to fine-grained spatial specifications~\cite{biagini2025hierasurg,chen2025surgsora,rapuri2026saw}, allowing more accurate specification of instrument motion. These advances are essential for building clinically useful simulators for controllable policy learning and safe evaluation.

Despite progress, existing SWMs still face two key challenges. 
\textbf{(i) Spatial Interaction Incoherence.} Existing works directly optimize pixel distribution without imposing explicit cross-frame motion constraints, often causing spatially inconsistent tissue responses under visible instrument contact (Fig.~\ref{fig:teaser}c). While some methods~\cite{chen2025surgsora,rapuri2026saw} incorporate geometric cues such as depth priors to mitigate this issue, tool-induced tissue deformation in surgical scenes is highly localized and morphologically complex, requiring fine-grained motion correspondence that global geometric priors cannot provide. \textbf{(ii) Temporal Fidelity Collapse.} Existing SWMs suffer from severe degradation in long-horizon simulation. Over rollouts spanning tens of seconds, generated frames progressively exhibit overexposure, color drift, and structural corruption (Fig.~\ref{fig:teaser}d), as prediction errors propagate through autoregressive rollouts~\cite{huang2025selfforcing,cui2025selfforcingpp}.
Complex tissue textures and challenging illumination further amplify this degradation, limiting the effectiveness of general-domain solutions~\cite{li2025svi,wang2026matrix}.

To address these issues, we present \textbf{SurgVista}, a surgical world model that achieves plausible instrument-tissue dynamics over long horizons, building upon two complementary training strategies. \emph{(i) Deformation Consistency Regularization} extracts dense scene-point trajectories that jointly cover instruments and tissue, and enforces contrastive coherence in the low-dimensional manifold, improving instrument-tissue motion consistency. 
\emph{(ii) Drift Adaptation Training} mitigates long-horizon drift by perturbing conditioning frames with online prediction residuals and rollout-calibrated photometric augmentations, sustaining visual fidelity over extended rollouts.
We further propose \textbf{SurgWorld-Bench}, the first benchmark featuring diverse surgical procedure types, long-range surgical operations, and hierarchical evaluation of instrument-motion accuracy and tissue-response fidelity. 
Extensive experiments show that SurgVista consistently outperforms state-of-the-art methods, with gains increasing as the horizon lengthens. Our contributions are as follows:

\begin{itemize}
    \item We present SurgVista, a surgical world model that mitigates spatial interaction incoherence and temporal fidelity collapse through two dedicated training recipes, achieving physically consistent long-horizon surgical scene simulation.
    \item We introduce Deformation Consistency Regularization, which enforces cross-frame coherence along scene-point trajectories, suppressing incoherent instrument-tissue dynamics.
    \item We introduce Drift Adaptation Training, which perturbs training conditions with online prediction residuals and photometric degradations, enabling longer-horizon stable rollouts.
    \item We present SurgWorld-Bench, a multi-horizon multi-procedure benchmark with decoupled metrics for instrument-motion accuracy and tissue-response fidelity.
\end{itemize}

\section{Related Work}
\label{sec:related}

\subsection{Surgical World Models}
Classical world models learn compact latent dynamics for policy optimization in simulated environments~\cite{ha2018world,hafner2020dreamer,hafner2023mastering}, while recent video diffusion models demonstrate that pixel-space generation itself can approximate a simulator~\cite{bruce2024genie,sun2025worldplay,he2025matrix}.
In surgery, existing world models formulate future prediction as pixel-wise video generation conditioned on varying action representations.
One line of work~\cite{he2025surgworld,turkcan2025towards,rapuri2026saw} fine-tunes foundation video DiTs~\cite{yang2024cogvideox,wu2025hunyuanvideo} on surgical footage, enabling surgical world modeling with improved realism.
In parallel, the granularity of action control has advanced from coarse textual descriptions~\cite{he2025surgworld,turkcan2025towards} to fine-grained spatial signals such as panoptic segmentation maps~\cite{biagini2025hierasurg}, optical flow~\cite{chen2025surgsora}, and affordance masks combined with tool-tip trajectories~\cite{rapuri2026saw}.
Recent efforts further incorporate depth priors as auxiliary inputs~\cite{chen2025surgsora} or supervision~\cite{rapuri2026saw} to improve geometric plausibility, yet cross-frame motion correspondence remains underexplored, leading to spatial interaction incoherence.
Moreover, existing methods are evaluated on short clips and suffer from temporal fidelity collapse over long-horizon predictions.
SurgVista targets both gaps with two dedicated training recipes: Deformation Consistency Regularization to enforce cross-frame interaction coherence, and Drift Adaptation Training to sustain fidelity over long horizons.

\subsection{Long-Horizon World Modeling}
Extending video generation models to long-horizon prediction typically relies on autoregressive conditioning, where each new clip is generated from preceding frames.
This introduces a train-inference distribution gap~\cite{ning2023elucidating,schmidt2019generalization}: the model sees clean ground-truth contexts during training yet conditions on its own imperfect predictions at inference, causing errors to accumulate over time.
Existing remedies fall into two categories.
The first reduces dependency on preceding frames through modified noise schedules~\cite{chen2024diffusion,song2025history}, anchor frames~\cite{henschel2025streamingt2v,wang2026matrix}, or anti-drifting sampling~\cite{zhang2025framepack}, alleviating but not eliminating drift.
The second exposes the model to degraded conditioning during training to build robustness.
Self-Forcing-style methods~\cite{huang2025selfforcing,cui2025selfforcingpp} unroll multi-step autoregressive rollouts within each iteration, but require expensive bidirectional-to-causal distillation.
Error recycling~\cite{li2025svi,wang2026matrix} offers a lighter alternative by re-injecting previously collected prediction errors into clean inputs, but the stored errors originate from different samples or outdated checkpoints, poorly approximating the degradation that the current model would actually produce.
Our Drift Adaptation Training sidesteps these limitations: a single forward step probes sample-specific degradation while keeping the simulated drift synchronized with the current model, and photometric augmentations further simulate pixel-level artifacts that latent-space error injection alone overlooks.
\section{SurgVista}
\textbf{Problem Formulation.} 
Given an observed frame $\mathbf{s}_0 \in \mathbb{R}^{H \times W \times 3}$ and an action sequence $\mathbf{a}_{1:T}$, SurgVista predicts future frames $\hat{\mathbf{s}}_{1:T}$ that capture both instrument motion and the induced tissue deformation.
For long-horizon simulation beyond a single clip, generated clips are chained autoregressively, where the tail frames of each generated clip condition the next.

\textbf{Overview.} 
SurgVista builds on a pretrained latent video DiT~\cite{wan2025wan}.
Let $\mathbf{z}_1 = \mathcal{E}(\mathbf{s}_{1:T}) \in \mathbb{R}^{T_z \times H_z \times W_z \times C}$ denote the target latent encoded by a pretrained VAE encoder $\mathcal{E}$, and let $\mathbf{z}_0 \sim \mathcal{N}(\mathbf{0}, \mathbf{I})$ be the noise sample.
Flow matching constructs the interpolant $\mathbf{z}_\tau = \tau \mathbf{z}_1 + (1-\tau)\mathbf{z}_0$ at flow time $\tau \sim \mathcal{U}[0,1]$ and trains the DiT to regress the velocity field $v_\theta(\mathbf{z}_\tau, \tau)$.
To condition generation on instrument actions, each action $\mathbf{a}_t$ is represented as a set of instrument keypoint coordinates, forming trajectories $\{\mathbf{p}^k_{1:T}\}_{k=1}^K$ that specify instrument positions over time.
Latent Action Encoding (Sec.~\ref{sec:lae}) converts these trajectories into an action latent $\mathbf{M}$ by replicating first-frame latent features at each trajectory's target positions.
The action latent $\mathbf{M}$ is concatenated with $\mathbf{z}_\tau$ along the channel dimension, while $\mathbf{z}_{\text{first}} = \mathcal{E}(\mathbf{s}_0)$ is projected through a reference convolution into tokens prepended to the DiT input sequence.
Text prompts are encoded and injected via cross-attention.
We denote all conditioning collectively as $\mathbf{c}$.
The model is trained by minimizing
\begin{equation}
    \mathcal{L}_{\text{FM}} = \mathbb{E}_{\tau,\,\mathbf{z}_0}\left[\left\| v_\theta(\mathbf{z}_\tau, \mathbf{c}, \tau) - (\mathbf{z}_1 - \mathbf{z}_0)\right\|^2\right].
    \label{eq:fm}
\end{equation}

This pipeline produces visually plausible short clips, yet two limitations remain.
The flow-matching objective does not enforce cross-frame motion coherence, leading to spatial interaction incoherence in instrument-tissue dynamics.
Autoregressive rollout further exposes the model to its own imperfect predictions, causing temporal fidelity collapse over long horizons.
SurgVista introduces two training strategies: \textbf{Deformation Consistency Regularization}~(DCR, Sec.~\ref{sec:dcr}), which enforces cross-frame motion coherence along tracked scene-point trajectories, and \textbf{Drift Adaptation Training}~(DAT, Sec.~\ref{sec:dat}), which closes the train-inference conditioning gap through online drift simulation.

\begin{figure}
    \centering
    \includegraphics[width=1.0\linewidth]{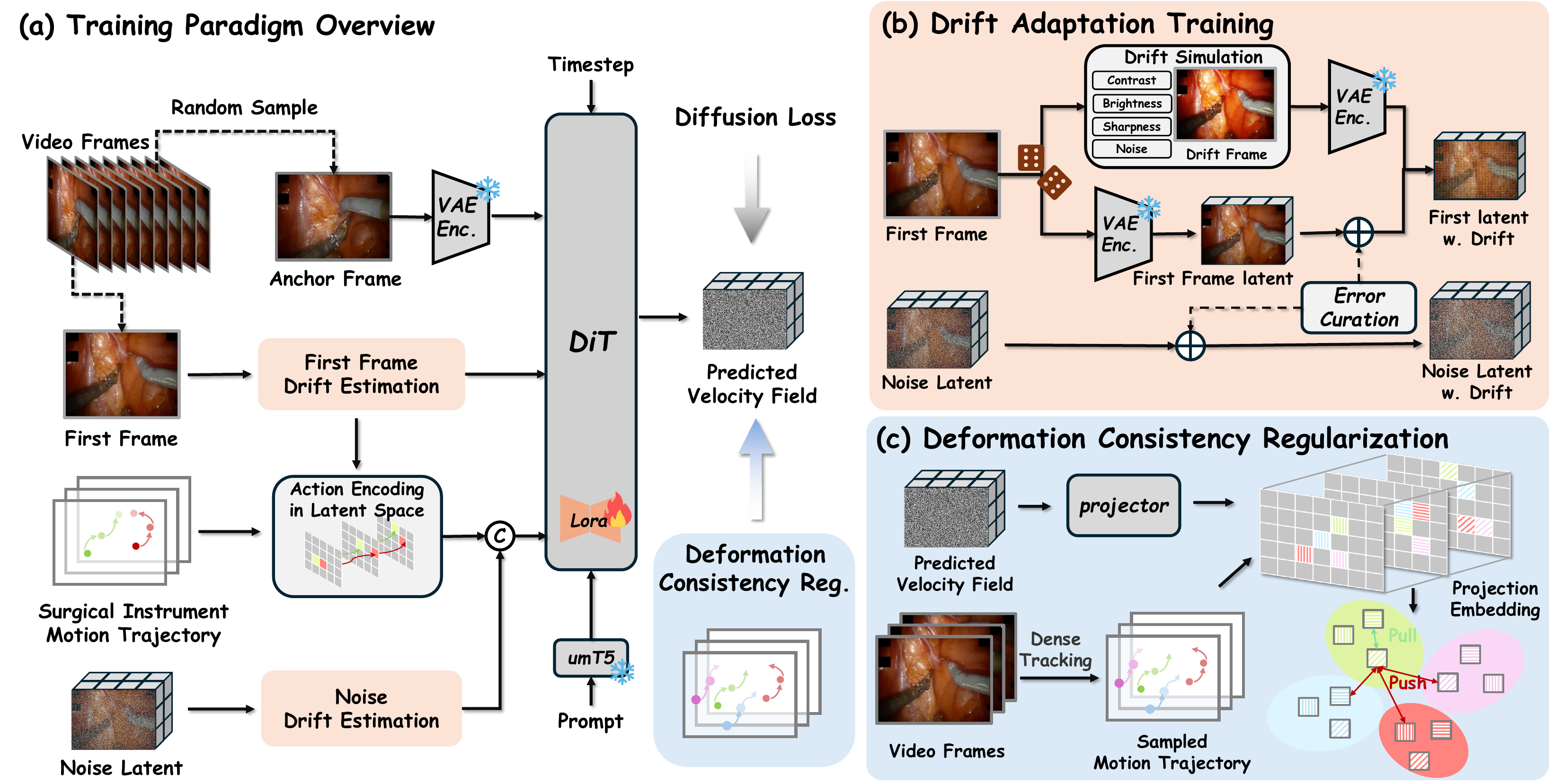}
\caption{\textbf{Overview of SurgVista.}
(a) Training paradigm, consisting of two training recipes: Deformation Consistency Regularization (DCR) and Drift Adaptation Training (DAT).
(b) DAT simulates inference-time conditioning drift via two complementary proxies.
(c) DCR enforces cross-frame motion coherence via contrastive learning on tracked scene-point trajectories.}
    \label{fig:framework}
\vspace{-10pt}
\end{figure}

\subsection{Latent Action Encoding}
\label{sec:lae}
Prior surgical world models encode instrument actions through segmentation masks~\cite{biagini2025hierasurg}, optical flow~\cite{chen2025surgsora}, or trajectory videos~\cite{rapuri2026saw}, requiring auxiliary models that increase computational overhead.
SurgVista instead converts instrument keypoint trajectories into an action latent by replicating first-frame latent features along each trajectory~\cite{chu2025wan}.
Concretely, given the initial-frame latent $\mathbf{z}_{\text{first}} = \mathcal{E}(\mathbf{s}_0) \in \mathbb{R}^{H_z \times W_z \times C}$ and pixel-space trajectories $\{\mathbf{p}^k_{1:T}\}_{k=1}^{K}$, each trajectory is mapped onto the spatiotemporal latent grid by downscaling spatial coordinates by the VAE spatial factor $f_s$ and grouping every $f_t$ consecutive frames into one latent frame, yielding latent-space trajectories $\{\hat{\mathbf{p}}^k_{1:T_z}\}_{k=1}^{K}$ with $T_z = T / f_t$.
The latent feature at each keypoint's initial position is then replicated to its target position in every subsequent latent frame, producing the action latent $\mathbf{M} \in \mathbb{R}^{T_z \times H_z \times W_z \times C}$:
\begin{equation}
    \mathbf{M}\!\left[\hat{t},\, \hat{\mathbf{p}}^k_{\hat{t}}\right]
    = \mathbf{z}_{\text{first}}\!\left[\hat{\mathbf{p}}^k_1\right], \quad \forall\, k,\, \hat{t} > 0,
    \label{eq:lae_rep}
\end{equation}
with all other entries set to zero.
The resulting $\mathbf{M}$ supplies spatially grounded action guidance without introducing any auxiliary modules.

\subsection{Deformation Consistency Regularization}
\label{sec:dcr}
Surgical world modeling demands motion coherence of instrument-tissue dynamics across frames, yet standard generative training solely models pixel distributions and provides no mechanism to enforce such correspondence.
As a result, generated sequences can exhibit interaction incoherence: an instrument contacts tissue, yet the surrounding region fails to produce a visually coherent deformation response.
To address this, we introduce Deformation Consistency Regularization (DCR), which imposes a motion-coherence constraint on the predicted video latent, as illustrated in Fig.~\ref{fig:framework}c.

Keypoint trajectories offer a unified motion description for both rigid instruments and deformable tissue, making them a natural supervision signal for cross-frame coherence.
We treat such trajectories as cross-frame correspondence anchors and enforce that the predicted latent assigns a consistent representation to each tracked point over time.
Unlike auxiliary losses that require decoding to pixel space, DCR operates in the latent space, avoiding the prohibitive memory cost of VAE decoding.

We employ an off-the-shelf point tracker~\cite{karaev2025cotracker3} to extract dense grid trajectories $\{(\mathbf{q}^j_t,\, o^j_t)\}_{t=1}^{T}$ for $j=1,\dots,J$ from each training video (details in Appendix C), where $\mathbf{q}^j_t \in \mathbb{R}^2$ denotes the image-space location of point $j$ at frame $t$ and $o^j_t \in [0,1]$ its visibility.
At each training step, given a sampled flow time $\tau\sim\mathcal{U}[0,1]$ with noisy latent $\mathbf{z}_\tau$, we obtain the predicted clean latent via a single Euler step,
\begin{equation}
    \hat{\mathbf{z}}_1
    = \mathbf{z}_\tau
    + (1-\tau)\cdot v_\theta\!\left(\mathbf{z}_\tau,\,\mathbf{c},\,\tau\right), 
    \label{eq:DCR_x1}
\end{equation}
and extract each tracked point's feature from the corresponding frame slice $\hat{\mathbf{z}}_{1,t}$ at latent-grid coordinate $\mathbf{q}^j_t / f_s$, projecting it through a trainable two-layer MLP $\phi$ into a $d$-dimensional embedding space,
\begin{equation}
    \mathbf{f}^j_t
    = \phi\!\left(
        \mathrm{sample}\!\left(\hat{\mathbf{z}}_{1,t},\,
        \mathbf{q}^j_t / f_s\right)
      \right),
    \label{eq:DCR_feat}
\end{equation}
yielding an $\ell_2$-normalized descriptor $\mathbf{f}^j_t\in\mathbb{R}^d$ at every visible frame $t\in\mathcal{O}^j=\{t:o^j_t>o_{\min}\}$.

For each trajectory $j$, a prototype $\bar{\mathbf{f}}^j=\frac{1}{|\mathcal{O}^j|}\sum_{t\in\mathcal{O}^j}\mathbf{f}^j_t$ aggregates descriptors over visible frames.
An InfoNCE loss~\cite{chen2020simple,he2020momentum} then pulls each descriptor toward its own prototype while pushing it away from the prototypes of the remaining $J{-}1$ trajectories:
\begin{equation}
    \mathcal{L}_{\text{DCR}}
    = -\frac{1}{|\mathcal{Q}|}
      \sum_{(j,t)\,\in\,\mathcal{Q}}
      \log \frac{
        \exp\!\left(\mathbf{f}^j_t \cdot \bar{\mathbf{f}}^j / \gamma\right)
      }{
        \displaystyle\sum_{j'=1}^{J}
        \exp\!\left(\mathbf{f}^j_t \cdot \bar{\mathbf{f}}^{j'} / \gamma\right)
      },
    \label{eq:DCR_loss}
\end{equation}
with $\mathcal{Q}=\{(j,t):t\in\mathcal{O}^j\}$ and temperature $\gamma$.
The positive term anchors each tracked point to a stable representation across time, while the negatives prevent trivial collapse.
DCR thus promotes both accurate instrument dynamics and realistic tissue deformation within each generated clip.

\subsection{Drift Adaptation Training}
\label{sec:dat}
When generation extends beyond a single clip through autoregressive conditioning, a train-inference distribution shift emerges: the model trains on clean ground-truth frames yet conditions on its own imperfect outputs at inference.
This mismatch compounds across clips and eventually dominates long-horizon failure.
The key idea of DAT is to deliberately corrupt conditioning frames during training, forcing the model to recover clean predictions from drifted inputs and thereby building robustness to imperfect inference-time conditions.
DAT replaces clean conditioning with corrupted surrogates through two complementary proxies (Fig.~\ref{fig:framework}b), one model-intrinsic and one model-agnostic.

\paragraph{Model-intrinsic proxy: online prediction residuals.}
An ideal training proxy would expose the model to perturbations drawn from the same error distribution it will encounter during rollout.
We achieve this by extracting the model's own prediction error from the training forward pass and injecting it back as a conditioning perturbation.

Specifically, we sample an independent flow time $\tau'\sim\mathcal{U}[0,1]$, construct $\mathbf{z}_{\tau'} = \tau'\mathbf{z}_1 + (1{-}\tau')\mathbf{z}_0'$ with a fresh noise sample $\mathbf{z}_0'\sim\mathcal{N}(\mathbf{0},\mathbf{I})$, and obtain the model's own prediction of the clean video latent via a single no-gradient Euler step,
\begin{equation}
    \hat{\mathbf{z}}_1'
    = \mathbf{z}_{\tau'}
    + (1{-}\tau')\cdot \mathrm{sg}\!\left[v_\theta\!\left(\mathbf{z}_{\tau'},\,\mathbf{c},\, \tau'\right)\right],
    \label{eq:pred_z1}
\end{equation}
where $\mathrm{sg}[\cdot]$ denotes stop-gradient and $\mathbf{c}$ uses the original clean conditioning.
The first temporal slice $\hat{\mathbf{z}}'_{1,0}$ is the model's prediction of the first frame.
Its deviation from the ground-truth conditioning latent yields the drift residual,
\begin{equation}
    \boldsymbol{\delta} = \hat{\mathbf{z}}'_{1,0} - \mathbf{z}_{\text{first}},
    \label{eq:residual}
\end{equation}
which is injected back onto the clean conditioning to simulate corrupted inference-time inputs,
\begin{equation}
    \tilde{\mathbf{z}}_{\text{first}}
    = \mathbf{z}_{\text{first}} + \boldsymbol{\delta}.
    \label{eq:inject}
\end{equation}
In addition, the full prediction error is injected into the noisy latent $\tilde{\mathbf{z}}_\tau = \mathbf{z}_\tau + (\hat{\mathbf{z}}_1' - \mathbf{z}_1)$, so that the denoising input jointly reflects the drift the model will encounter at inference.
Since $\boldsymbol{\delta}$ originates from the model's own imperfect prediction, it naturally mirrors the error characteristics the model will produce during autoregressive rollout.
Sampling $\tau'$ and $\mathbf{z}_0'$ independently of the main loss decouples the corruption severity from the training noise level, increasing the diversity of simulated drift.
We detach $\boldsymbol{\delta}$ from the computational graph but recompute it at every iteration, so the proxy co-evolves with $v_\theta$ without affecting its gradient.

\begin{wrapfigure}{r}{0.5\textwidth}
    \centering
    \includegraphics[width=0.48\textwidth]{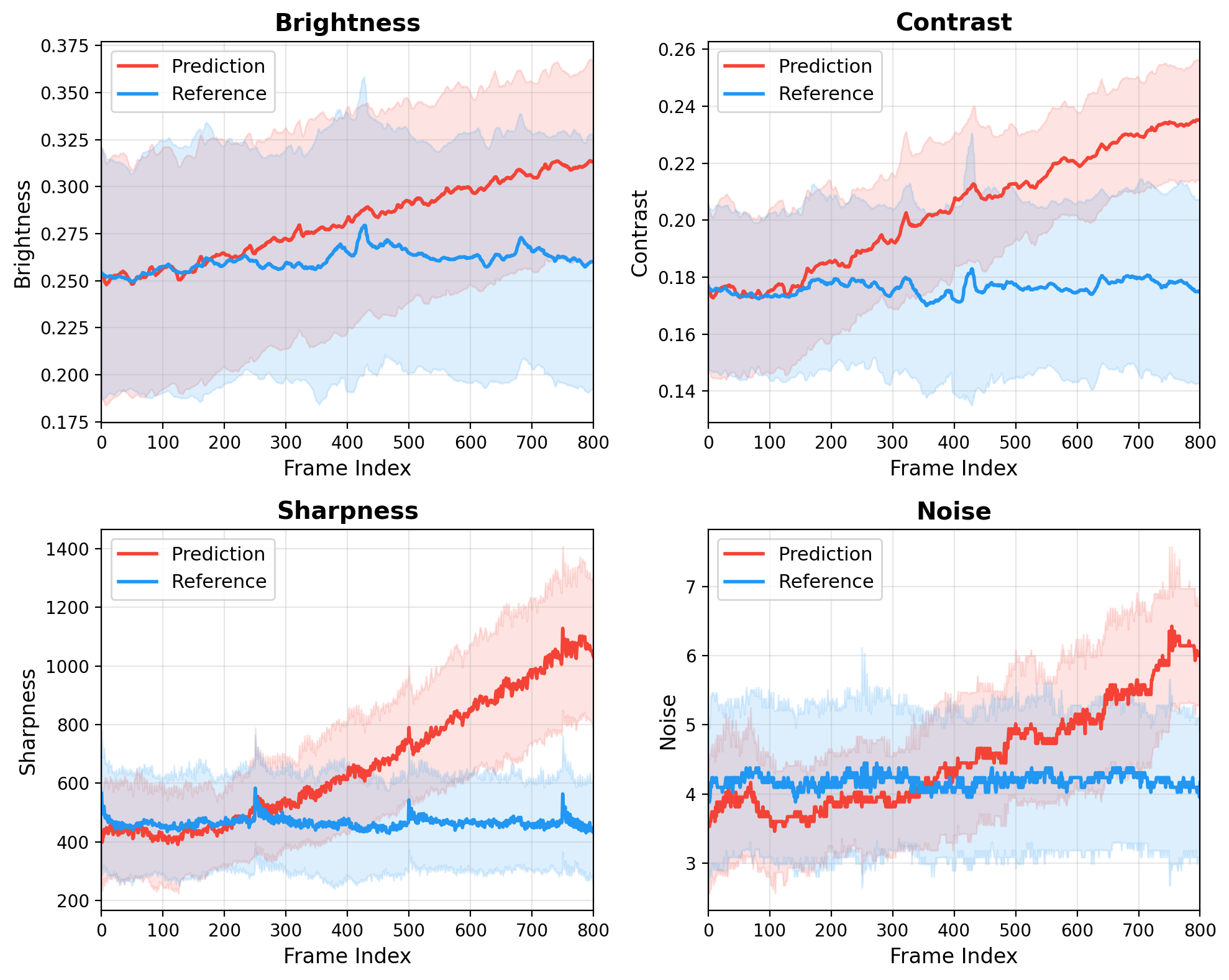}
    \vspace{-7pt}
    \caption{Main photometric statistics over long-horizon predictions.}
    \label{fig:drift_metrics}
    \vspace{-5pt}
\end{wrapfigure}
\paragraph{Model-agnostic proxy: photometric augmentations.}
As training progresses, the probe residual $\boldsymbol{\delta}$ shrinks with improving model accuracy.
Yet repeated decode-encode cycles during autoregressive rollout still introduce persistent photometric degradations, including contrast drift, color skew, texture blur, and specular blowout.
To quantify this effect, we measure photometric statistics over long-horizon rollouts and find that four key dimensions (brightness, contrast, sharpness, and noise level) diverge progressively from ground truth as the horizon grows (Fig.~\ref{fig:drift_metrics}).
Motivated by this observation, we randomly apply four photometric perturbations to the conditioning frame: exposure shift, contrast jitter, sharpness scaling, and noise adjustment, each with randomly sampled strength calibrated to match observed rollout artifacts.
The corrupted conditioning latent is then obtained as
\begin{equation}
    \tilde{\mathbf{z}}_{\text{first}}
    = \mathcal{E}\!\left(\mathcal{T}_{\text{img}}\!\left(\mathbf{s}_{\text{first}}\right)\right),
    \label{eq:photo_aug}
\end{equation}
where $\mathcal{T}_{\text{img}}$ denotes the sampled augmentation.
Since $\mathcal{T}_{\text{img}}$ is independent of $v_\theta$, this proxy provides a persistent source of corrupted conditioning regardless of the model's current prediction quality.

\paragraph{Training objective.}
At each training step, we randomly select one of three conditioning modes: the model-intrinsic proxy (Eq.~\ref{eq:inject}), the model-agnostic proxy (Eq.~\ref{eq:photo_aug}), or the clean latent $\mathbf{z}_{\text{first}}$ to preserve accuracy on uncorrupted inputs, with probabilities $p_{\text{probe}}$, $p_{\text{aug}}$, and $p_{\text{clean}}$ respectively.
The target remains the clean ground-truth clip regardless of the conditioning mode.
The flow-matching loss under the selected conditioning is
\begin{equation}
    \mathcal{L}_{\text{DAT}}
    = \mathbb{E}_{\tau,\,\mathbf{z}_0}\!\left[
        \left\|
            v_\theta\!\left(\tilde{\mathbf{z}}_\tau,\,
                \tilde{\mathbf{c}},\, \tau\right) 
            - \left(\mathbf{z}_1 - \mathbf{z}_0\right)
        \right\|^2
    \right],
    \label{eq:dat_loss}
\end{equation}
where $\tilde{\mathbf{c}}$ denotes the conditioning with $\mathbf{z}_{\text{first}}$
replaced by the selected corrupted variant $\tilde{\mathbf{z}}_{\text{first}}$
and supplemented with an anchor frame $\mathbf{s}_{\text{anchor}}$ fixed to the first frame throughout inference as a clean visual reference, and $\tilde{\mathbf{z}}_\tau$ equals $\mathbf{z}_\tau + (\hat{\mathbf{z}}_1' - \mathbf{z}_1)$ under the model-intrinsic proxy and reduces to the unmodified $\mathbf{z}_\tau$ under the other two modes.

\paragraph{Remark.}
Recent error-recycling methods~\cite{li2025svi,wang2026matrix} also target the train-inference gap, but draw corrupted conditioning from heterogeneous sources such as different samples or earlier checkpoints that may not reflect the current model's error distribution.
DAT instead computes the residual online from the same sample in each forward pass, so the simulated drift co-evolves with the model.
The photometric proxy further covers pixel-level artifacts that latent-space residuals alone cannot capture.

Combining the two training recipes, the full objective of SurgVista is
\begin{equation}
    \mathcal{L}
    = \mathcal{L}_{\text{DAT}}
    + \lambda_{\text{DCR}}\,\mathcal{L}_{\text{DCR}},
    \label{eq:total}
\end{equation}
where $\mathcal{L}_{\text{DAT}}$ drives temporal fidelity under conditioning drift and $\mathcal{L}_{\text{DCR}}$ enforces cross-frame motion coherence within each generated clip.
\section{SurgWorld-Bench}
\label{sec:benchmark}

\begin{figure}[h]
    \centering
    \includegraphics[width=1.0\linewidth]{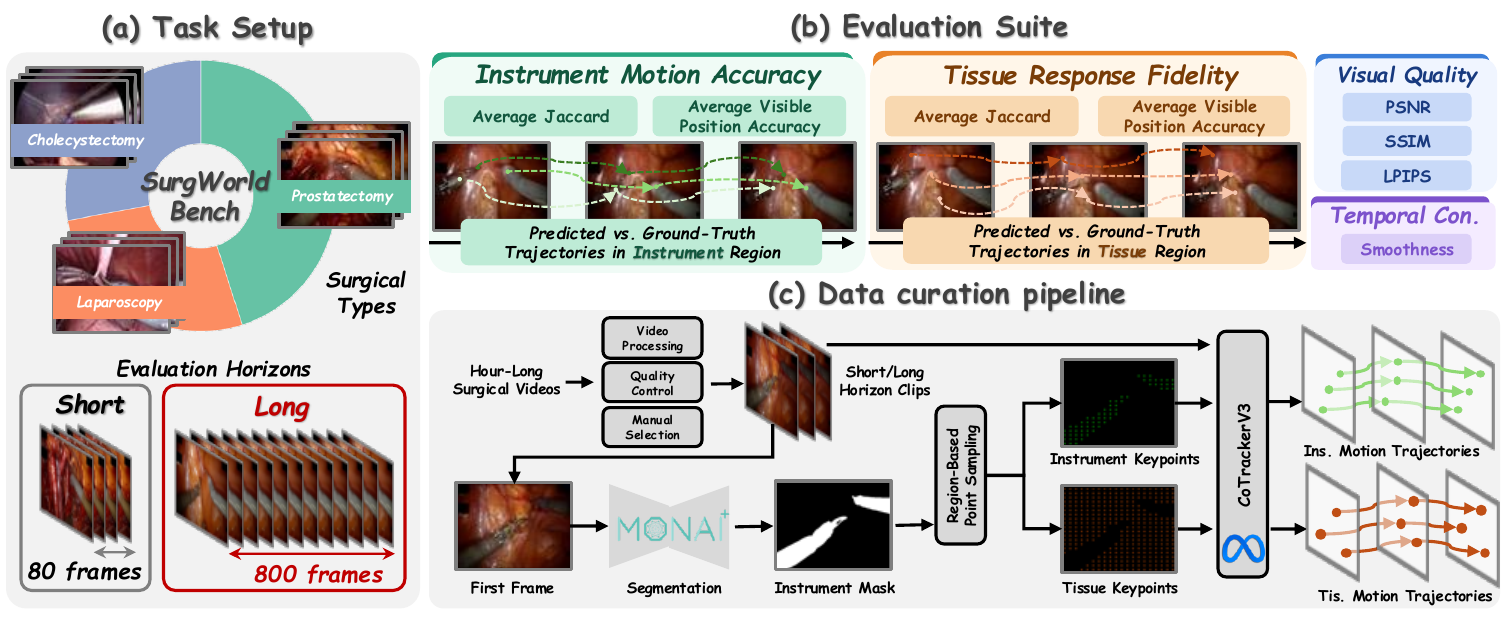}
    \vspace{-20pt}
    \caption{\textbf{Overview of SurgWorld-Bench.}
    (a)~Task setup with three procedure types and two evaluation horizons.
    (b)~Evaluation suite with decoupled instrument and tissue trajectory metrics.
    (c)~Data curation pipeline for extracting region-specific motion trajectories.}
    \label{fig:bench}
    \vspace{-10pt}
\end{figure}

Existing evaluations of surgical world models share three limitations: coverage of only a single procedure type, rollout lengths rarely exceeding 80 frames~\cite{he2025surgworld,biagini2025hierasurg,chen2025surgsora,rapuri2026saw}, and reliance on holistic metrics that conflate instrument kinematics with tissue biomechanics.
We introduce \textbf{SurgWorld-Bench}, a surgical world-model benchmark to \emph{jointly} address all three through (a)~\emph{procedural diversity}, (b)~\emph{long-horizon evaluation} up to $10{\times}$ the typical 80-frame rollout, and (c)~a \emph{decoupled evaluation protocol} that separately probes instrument motion and tissue response (Fig.~\ref{fig:bench}).
\paragraph{Procedural diversity.} SurgWorld-Bench draws from three public datasets covering three surgical procedure types (Fig.~\ref{fig:bench}a): GraSP~\cite{ayobi2024pixelwise} for radical prostatectomy, SurgToolLoc22~\cite{zia2023intuitive} for porcine laparoscopic procedures, and CholecTrack20~\cite{nwoye2025cholectrack20} for laparoscopic cholecystectomy, collectively encompassing diverse subjects, heterogeneous instrument types, and varied tissue biomechanics.
\paragraph{Task and evaluation horizons.}
Given an initial frame $\mathbf{s}_0$ and an instrument trajectory $\mathbf{a}_{1:T}$, the model predicts the future state sequence $\hat{\mathbf{s}}_{1:T}$.
The benchmark operates at two evaluation horizons (Fig.~\ref{fig:bench}a): a short horizon of 80 frames over 111 evaluation clips for comparability with prior work, and a long horizon of 800 frames over 21 evaluation clips to probe long-horizon stability.
\paragraph{Evaluation protocol.}
Holistic metrics (e.g., PSNR) alone cannot distinguish whether errors originate from instrument motion or tissue response.
SurgWorld-Bench therefore assesses predictions along four complementary dimensions (Fig.~\ref{fig:bench}b), where the first two retain the standard global view and the latter two supply region-decoupled diagnosis.
\textbf{(i)~Visual quality}: per-frame PSNR, SSIM, and LPIPS~\cite{zhang2018unreasonable}.
\textbf{(ii)~Temporal consistency}: the Motion Smoothness Score~\cite{huang2024vbench}, which captures temporal coherence.
For \textbf{(iii)~Instrument-motion accuracy} and \textbf{(iv)~Tissue-response fidelity}, each frame is partitioned into instrument and tissue regions by a segmentation model~\cite{cardoso2022monai}, and a point tracker~\cite{karaev2025cotracker3} extracts densely sampled trajectories within each region on both predicted and ground-truth sequences (Fig.~\ref{fig:bench}c).
Two tracking metrics are reported for each region: Average Jaccard (AJ)~\cite{doersch2022tap} and Average Visible Position Accuracy (${\delta}_{avg}^{vis}$)~\cite{doersch2022tap}.
The instrument-region scores measure whether predicted instruments follow the prescribed action $\mathbf{a}_{1:T}$, while the tissue-region scores measure whether the induced deformation matches the ground-truth interaction.
Details and the validity analysis of the region-decoupled metrics are provided in Appendix B.
\section{Experiments}
\label{sec:exp}

\subsection{Setup}
\label{sec:setup}

SurgVista builds on Wan2.1-1.3B and is trained for 100K iterations at $512{\times}320$ resolution with a clip length of 81 frames.
Baselines fall into two groups: general-purpose video generation models conditioned on text-described actions (SVD~\cite{blattmann2023videoldm}, CogVideoX1.5-5B~\cite{yang2024cogvideox}, HunyuanVideo-1.5~\cite{wu2025hunyuanvideo}, Wan2.2-TI2V-5B~\cite{wan2025wan}), and controllable video generation models conditioned on dense action signals such as optical flow or point trajectories (Wan2.1-Control~\cite{wan2fun2025}, SurgSora~\cite{chen2025surgsora}).
All methods are evaluated on SurgWorld-Bench under short-horizon and long-horizon settings.
Dataset sources, training configurations, and additional implementation details are provided in Appendix C.

\begin{table*}[t]
\centering
\small
\setlength{\tabcolsep}{4pt}
\renewcommand{\arraystretch}{1.15}
\caption{\textbf{Comparison with state-of-the-art methods on SurgWorld-Bench} across instrument-motion accuracy, tissue-response fidelity, visual quality, and temporal consistency at short-horizon and long-horizon settings, with best results per column in \textbf{bold}.}
\label{tab:long_video_generation_comparison}
\begin{tabular}{clcccccccc}
\toprule
& \multirow{2}{*}{\textbf{Method}}
& \multicolumn{2}{c}{\shortstack{\textbf{Instrument}\\\textbf{Motion}}}
& \multicolumn{2}{c}{\shortstack{\textbf{Tissue}\\\textbf{Response}}}
& \multicolumn{3}{c}{\shortstack{\textbf{Visual}\\\textbf{Quality}}}
& \multicolumn{1}{c}{\shortstack{\textbf{Temporal}\\\textbf{Consistency}}} \\
\cmidrule(lr){3-4} \cmidrule(lr){5-6} \cmidrule(lr){7-9} \cmidrule(lr){10-10}
& & AJ $\uparrow$ & ${\delta}_{avg}^{vis}$ $\uparrow$
& AJ $\uparrow$ & ${\delta}_{avg}^{vis}$ $\uparrow$
& PSNR $\uparrow$ & SSIM $\uparrow$ & LPIPS $\downarrow$
& Smoothness $\uparrow$  \\
\midrule
\multirow{7}{*}{\rotatebox{90}{\textbf{Short}}}
& SVD~\cite{blattmann2023videoldm}                & 3.62  & 7.19  & 8.69  & 16.22 & 13.80 & 0.42 & 0.48 & 96.61 \\
& CogVideoX1.5-5B~\cite{yang2024cogvideox}    & 4.34  & 8.24  & 15.21 & 26.46 & 13.14 & 0.42 & 0.46 & 96.91 \\
& HunyuanVideo-1.5~\cite{wu2025hunyuanvideo}   & 4.80  & 9.89  & 18.93 & 30.91 & 16.35 & 0.51 & 0.36 & \textbf{99.14} \\
& Wan2.2-TI2V-5B~\cite{wan2025wan}     & 5.83  & 10.40 & 19.93 & 32.32 & 15.63 & 0.51 & 0.40 & 98.77 \\
& Wan2.1-Control~\cite{wan2fun2025}     & 39.34 & 52.08 & 36.19 & 49.51 & 19.08 & 0.62 & 0.26 & 99.03 \\
& SurgSora~\cite{chen2025surgsora}           & 19.41 & 27.00 & 19.51 & 30.51 & 15.12 & 0.43 & 0.49 & 98.50 \\
& \textbf{SurgVista (Ours)} & \textbf{45.76} & \textbf{58.43} & \textbf{41.68} & \textbf{55.29} & \textbf{19.85} & \textbf{0.64} & \textbf{0.24} & \textbf{99.14}  \\
\midrule
\multirow{7}{*}{\rotatebox{90}{\textbf{Long}}}
& SVD~\cite{blattmann2023videoldm}                & 1.61  & 2.46  & 3.55  & 5.54  & 7.99  & 0.23 & 0.76 & 96.26 \\
& CogVideoX1.5-5B~\cite{yang2024cogvideox}    & 2.69  & 4.06  & 7.65  & 11.41 & 10.44 & 0.39 & 0.67 & 96.75 \\
& HunyuanVideo-1.5~\cite{wu2025hunyuanvideo}   & 1.56  & 3.73  & 5.63  & 11.76 & 14.24 & 0.41 & 0.57 & 99.07 \\
& Wan2.2-TI2V-5B~\cite{wan2025wan}     & 1.94  & 3.15  & 5.91  & 9.57  & 11.87 & 0.39 & 0.65 & 98.74 \\
& Wan2.1-Control~\cite{wan2fun2025}     & 18.92 & 28.30 & 19.91 & 31.90 & 16.35 & 0.50 & 0.42 & 99.51 \\
& SurgSora~\cite{chen2025surgsora}           & 5.36  & 7.57  & 6.08  & 9.37  & 8.95 & 0.28 & 0.79 & 98.55 \\
& \textbf{SurgVista (Ours)} & \textbf{27.13} & \textbf{39.61} & \textbf{24.15} & \textbf{38.19} & \textbf{17.35} & \textbf{0.55} & \textbf{0.37} & \textbf{99.53}  \\
\bottomrule
\end{tabular}
\vspace{-15pt}
\end{table*}

\subsection{Comparison with State-of-the-Art Methods}
\label{sec:comparison}

\paragraph{Quantitative results.}
Table~\ref{tab:long_video_generation_comparison} reports the main comparison on SurgWorld-Bench.
SurgVista outperforms all baselines across every evaluation dimension, and its advantage widens as the prediction horizon increases.
For short-horizon world modeling, general-purpose video generation models produce visually realistic frames yet lack fine-grained controllability over instrument motion, resulting in instrument motions that deviate from the specified actions.
SurgSora, designed specifically for surgical video modeling, exhibits stronger instrument control but is constrained by its backbone capacity, trailing SurgVista in tissue-response fidelity and visual quality.
The cross-frame motion supervision of DCR enables SurgVista to achieve more accurate instrument control and more realistic tissue responses while maintaining superior visual quality.
Long-horizon world modeling poses a greater challenge, as all methods suffer degradation in visual quality, instrument-control accuracy, and tissue-response realism.
By simulating distribution drift during training, the DAT mechanism markedly improves rollout stability, allowing SurgVista to maintain a clear margin over all baselines.

\begin{figure}[h]
    \centering
    \vspace{-8pt}
    \includegraphics[width=1.0\linewidth]{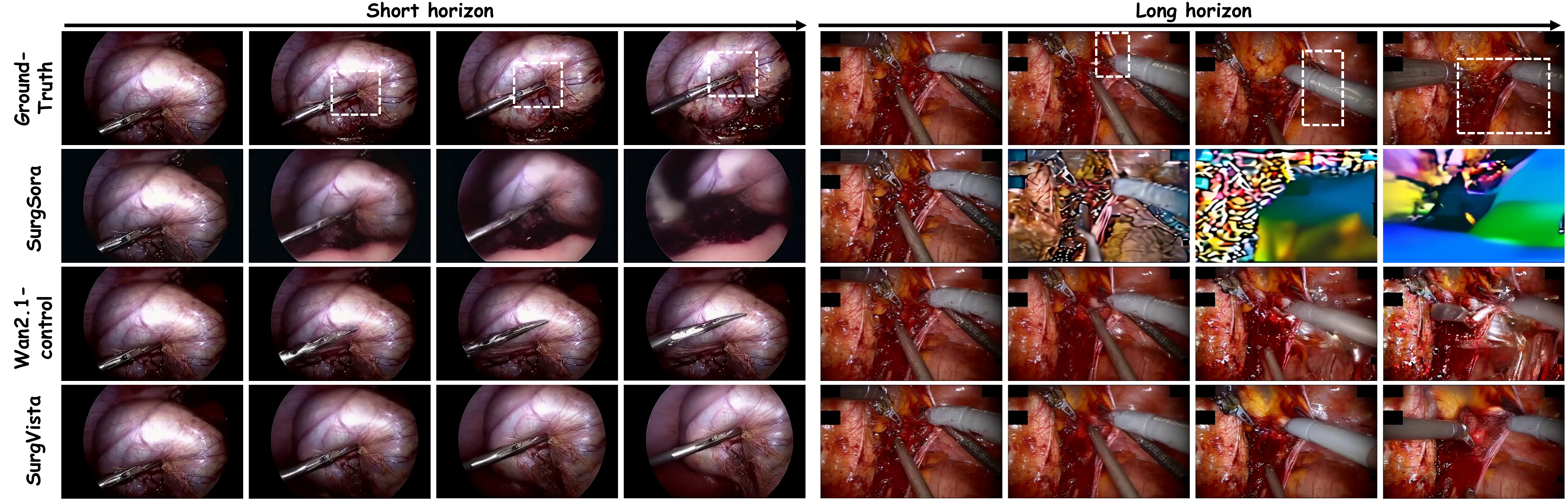}
    \vspace{-15pt}
\caption{\textbf{Qualitative comparison with state-of-the-art methods.}
    Left: short-horizon predictions at frames 1, 5, 10, and 15. Right: long-horizon predictions at frames 1, 100, 500, and 800.}
    \label{fig:qualitative}
\end{figure}

\paragraph{Qualitative comparison.}
Figure~\ref{fig:qualitative} presents a visual comparison of SurgVista against SurgSora and Wan2.1-Control under both short- and long-horizon settings.
In the short-horizon setting, SurgSora exhibits inaccurate instrument motion accompanied by noticeable visual artifacts. Wan2.1-Control reproduces instrument motion faithfully, yet the surrounding tissue remains largely static, missing the deformations expected upon instrument contact. SurgVista captures both accurate instrument motion and the resulting tissue response, with highlighted regions revealing realistic deformation around the interaction point.
Over longer horizons, SurgSora nearly collapses, while Wan2.1-Control accumulates color drift and structural blur.
By contrast, SurgVista maintains more realistic appearance, faithful instrument geometry, and plausible tissue responses throughout the rollout.
Additional qualitative results are provided in Appendix A.

\subsection{Ablation Studies}
\label{sec:ablation}

Table~\ref{tab:ablation} ablates the contributions of DCR and DAT under long-horizon prediction.
DCR alone improves both instrument-motion and tissue-deformation accuracy through cross-frame motion regularization.
DAT alone yields larger gains by simulating conditioning drift during training, directly targeting the autoregressive distribution gap that dominates long-horizon failure.
Combining both achieves the best results across all metrics, indicating that the spatial coherence provided by DCR and the drift robustness provided by DAT address distinct aspects of long-horizon degradation.

Table~\ref{tab:simulation} compares three drift simulation strategies, all sharing DCR and the anchor frame: clean-only training, error recycling from~\cite{li2025svi}, and the full DAT design.
The anchor alone brings only modest gains.
Error recycling~\cite{li2025svi} reuses prediction residuals from other samples and earlier checkpoints, offering limited improvement.
The online self-prediction strategy in DAT instead recomputes per-sample residuals under the current model weights, yielding drift signals better adapted to the current training state and stronger overall performance. 
Adding image-space drift simulation further captures pixel-level drift characteristics, yielding the best overall performance.
\begin{table}[t]
\centering
\small
\setlength{\tabcolsep}{3pt}
\renewcommand{\arraystretch}{1.15}
\begin{minipage}[t]{0.49\linewidth}
\vspace{0pt}
\centering
\caption{\textbf{Component ablation} on instrument and tissue accuracy (long-horizon).}
\vspace{-5pt}
\label{tab:ablation}
\begin{tabular}{l cc cc}
\toprule
& \multicolumn{2}{c}{Ins.} & \multicolumn{2}{c}{Tis.} \\
\cmidrule(lr){2-3} \cmidrule(lr){4-5}
Method & AJ & ${\delta}_{avg}^{vis}$ & AJ & ${\delta}_{avg}^{vis}$ \\
\midrule
Baseline            & 19.45 & 29.09 & 20.00 & 32.20  \\
\; + DCR only       & 20.57 & 31.10 & 21.18 & 33.67 \\
\; + DAT only           &  26.54 & 38.83 & 24.10 & 37.96 \\
\; + DCR + DAT (full)         & \textbf{27.13} & \textbf{39.61} & \textbf{24.15} & \textbf{38.19}  \\
\bottomrule
\end{tabular}
\end{minipage}
\hfill
\begin{minipage}[t]{0.49\linewidth}
\vspace{0pt}
\centering
\caption{\textbf{Effect of drift simulation strategies} on instrument and tissue accuracy (long-horizon).}
\vspace{-5pt}
\label{tab:simulation}
\begin{tabular}{l cc cc}
\toprule
& \multicolumn{2}{c}{Ins.} & \multicolumn{2}{c}{Tis.} \\
\cmidrule(lr){2-3} \cmidrule(lr){4-5}
Method & AJ & ${\delta}_{avg}^{vis}$ & AJ & ${\delta}_{avg}^{vis}$ \\
\midrule
Clean only  & 22.11 & 33.64 & 21.87 & 35.16  \\
Error recycling~\cite{li2025svi} & 24.15 & 35.96 & 23.35 & 36.76\\
DAT -- latent          & 26.09 & 38.55 & 24.09 & 37.74 \\
DAT -- latent + image & \textbf{27.13} & \textbf{39.61} & \textbf{24.15} & \textbf{38.19} \\
\bottomrule
\end{tabular}
\end{minipage}
\vspace{-5pt}
\end{table}

\subsection{Discussion}
\label{sec:discussion}

\begin{table}[t]
\centering
\small
\setlength{\tabcolsep}{3pt}
\renewcommand{\arraystretch}{1.15}
\begin{minipage}[t]{0.49\linewidth}
\vspace{0pt}
\centering
\caption{\textbf{Effect of DCR spatial region} on instrument and tissue accuracy (short-horizon).}
\vspace{-5pt}
\label{tab:DCR_region}
\begin{tabular}{l cc cc}
\toprule
& \multicolumn{2}{c}{Ins.} & \multicolumn{2}{c}{Tis.} \\
\cmidrule(lr){2-3} \cmidrule(lr){4-5}
Method & AJ & ${\delta}_{avg}^{vis}$ & AJ & ${\delta}_{avg}^{vis}$ \\
\midrule
Baseline            & 40.42  & 53.13 & 36.07 & 49.43  \\
DCR -- Instrument only       & 44.91 & 57.62 & 38.39 & 52.29 \\
DCR -- Tissue only           & 44.42 & 57.21 & 41.23 & 54.90 \\
DCR -- All (default)         & \textbf{45.76} & \textbf{58.43} & \textbf{41.68} & \textbf{55.29}  \\
\bottomrule
\end{tabular}
\end{minipage}
\hfill
\begin{minipage}[t]{0.49\linewidth}
\vspace{0pt}
\centering
\caption{\textbf{Effect of DCR trajectory count} on instrument and tissue accuracy (short-horizon).}
\vspace{-5pt}
\label{tab:DCR_grid}
\begin{tabular}{l cc cc}
\toprule
& \multicolumn{2}{c}{Ins.} & \multicolumn{2}{c}{Tis.} \\
\cmidrule(lr){2-3} \cmidrule(lr){4-5}
Method & AJ & ${\delta}_{avg}^{vis}$ & AJ & ${\delta}_{avg}^{vis}$ \\
\midrule
Baseline              & 40.42  & 53.13 & 36.07 & 49.43 \\
DCR -- $32$          & 44.65 & 57.25 & 41.07 & 54.68 \\
DCR -- $64$ (default) & \textbf{45.76} & \textbf{58.43} & 41.68 & 55.29 \\
DCR -- $128$        & 45.74 & 58.37 & \textbf{41.70} &  \textbf{55.37} \\
\bottomrule
\end{tabular}
\end{minipage}
\vspace{-12pt}
\end{table}

\textit{Effect of DCR on different regions.}
To investigate how the supervision region affects DCR, the constraint is applied to the instrument, tissue, or full-frame region separately.
As Table~\ref{tab:DCR_region} shows, instrument-only supervision maximizes instrument metrics, tissue-only supervision maximizes tissue metrics, and full-frame supervision yields the most balanced improvement.
% We therefore adopt full-frame supervision by default, as a practical surgical world model demands both accurate instrument control and realistic tissue response.

\textit{Effect of trajectory count in DCR.}
Table~\ref{tab:DCR_grid} varies the number of sampled trajectories.
Increasing from 32 to 64 yields consistent improvements, as denser sampling provides richer coverage of instrument and tissue dynamics, while further increasing to 128 leads to performance saturation.
We therefore adopt 64 as the default, balancing supervision quality and computational overhead.

\textit{Failure cases.}
SurgVista still struggles with large-scale tissue deformations and extremely fine-grained manipulations. A detailed analysis of these failure modes is provided in Appendix D.

\section{Conclusion}
\label{sec:conclusion}
This paper presents SurgVista, a surgical world model that sustains plausible instrument-tissue interaction over long-horizon predictions. Deformation Consistency Regularization enforces cross-frame motion coherence along dense scene-point trajectories to improve interaction plausibility, while Drift Adaptation Training perturbs conditioning frames with online prediction residuals and photometric augmentations to sustain visual fidelity over extended rollouts. SurgWorld-Bench provides a multi-procedure, multi-horizon benchmark with decoupled metrics. Experiments confirm consistent gains over existing methods, with advantages amplifying at longer horizons. Extending the action space to incorporate force and tactile modalities is a promising future direction.

{
    \small
    \bibliographystyle{rusnat}
    \bibliography{reference}
}

\end{document}